\documentclass[11pt]{article}
\newif\ifcomment\commentfalse

\usepackage{color}
\usepackage{graphicx}
\usepackage{url}
\usepackage{latexsym}

\usepackage{nomencl}

\newcommand{\email}[1]{\begin{small} \url{#1} \end{small}}
\newcommand{\smallurl}[1]{ \begin{scriptsize}\url{#1}\end{scriptsize}}

\definecolor{grey}{rgb}{0.95,0.95,0.95}
\usepackage{style/acl2015}
\usepackage{times}

\makeatletter
\newcommand{\@BIBLABEL}{\@emptybiblabel}
\newcommand{\@emptybiblabel}[1]{
\makeatother}
\usepackage[draft]{hyperref}

\usepackage[bf,small]{caption}
\usepackage{subcaption}

\usepackage{booktabs}

\usepackage{graphicx}
\usepackage{latexsym}
\usepackage{xspace}
\usepackage[table]{xcolor}

\newcommand{\xhdr}[1]{\vspace{1.7mm}\noindent{{\bf #1.}}}

\newcommand{\betrayer}{betrayer\xspace}
\newcommand{\victim}{victim\xspace}
\newcommand{\B}{B\xspace}
\newcommand{\V}{V\xspace}
\newcommand{\planning}{planning\xspace}
\newcommand{\Planning}{Planning\xspace}

\newcommand{\nobetrayals}{lasting friendships\xspace}

\newcommand{\sectionrule}{\addlinespace[1ex]}

\title{Linguistic Harbingers of Betrayal:\\ A Case Study on an Online Strategy Game}

\author{
	Vlad Niculae,$^{1}$ Srijan Kumar,$^{2}$ Jordan
        Boyd-Graber,$^{3}$ Cristian Danescu-Niculescu-Mizil$^{1}$\\
        $^1$Cornell University, $^2$University of Maryland, $^3$University of Colorado\\
  	\small {\tt \href{mailto:vlad@cs.cornell.edu}{vlad@cs.cornell.edu}},
	{\tt \href{mailto:srijan@cs.umd.edu}{srijan@cs.umd.edu}}, \\
	\small {\tt \href{mailto:Jordan.Boyd.Graber@colorado.edu}{Jordan.Boyd.Graber@colorado.edu}},
	{\tt \href{mailto:cristian@cs.cornell.edu}{cristian@cs.cornell.edu}} \\
}

\begin{document}

\maketitle


\begin{abstract}

  %
  %
  %
  %
  Interpersonal relations are
   fickle, with close friendships often
  dissolving into enmity.  In this work, we explore linguistic cues that presage
  such transitions by studying dyadic interactions in
   an online strategy game where players form alliances and break those alliances through
  betrayal.  We characterize friendships that are unlikely to last and examine
  temporal patterns that foretell
  betrayal.

  We reveal that subtle signs of imminent betrayal are encoded in the
  conversational patterns of the dyad, even
  if the victim
  is not aware
  of the relationship's fate.  In particular, we find that lasting friendships
  exhibit a form of
balance
    that manifests itself through
  language.  In contrast, sudden changes in the balance of certain
  conversational
  attributes---such as positive sentiment, politeness, or focus on
  future planning---signal impending betrayal.

\end{abstract}


\section{Introduction}\label{sec:intro}

A major focus in computational social science has
been
 the study of interpersonal relations
  through data.
However, social interactions are complicated,
and
 we rarely
 have access
  all of the data that
define
 the relationship
between friends or enemies.
As an alternative, thought experiments like the prisoner's
dilemma~\cite{axelrod-88} are used to explain behavior.  Two prisoners---denied
communication---must decide whether to cooperate with each other or defect.
Such simple and elegant tools
initially
helped understand many real world scenarios from
pricing products~\cite{rosenthal-81} to athletes doping~\cite{buechel-13}.
Despite its power, the prisoner's dilemma remains woefully unrealistic.
Cooperation and betrayal do not happen in a cell cut off from the rest of the
world.  Instead, real interactions are mediated by communication: promises are
made, then broken, and met with recriminations.

To study the complex social phenomenon of betrayal, we turn to data
and observe the players of
{\bf Diplomacy} \cite{sharp1978game},
a war-themed strategy game
where
friendships
 and betrayals are orchestrated primarily through language.
Diplomacy, like the prisoner's
dilemma, is a repeated game where players choose to either cooperate or betray
other players.  Diplomacy is so engaging that it is played around the world,
not only casually as a board game but also
over the Internet
and
in formal
settings such as world championships.\footnote{\footnotesize A recent episode of \textit{This American Life} describes the Diplomacy game in a competitive offline setting: \small
\url{http://www.thisamericanlife.org/radio-archives/episode/531/got-your-back?act=1}}
Players converse throughout the game
and
victory hinges on enlisting others' support through persuasiveness and
cunning duplicity.
To illustrate the social relations that carry out throughout the game, consider the following
exchange between two
Diplomacy allies:
\begin{quote}
\small
Germany: Can I suggest you move your armies east and then I will support you?
Then next year you move {\em [there]} and dismantle Turkey.
I will deal with England and France, you take out Italy.

Austria: Sounds like a perfect plan!
Happy to follow through.
And---thank you Bruder!

\end{quote}

Austria is very polite and positive in its reply, and appreciates Germany's support and
generosity.
They have been good allies for the better part of the game.
However, immediately after this exchange, Austria
suddenly invades German territory.
The intention to do so was so well concealed
that Germany did not see
the betrayal
coming; otherwise
it
would have taken advantage
first. Indeed, if we follow their conversation after
the attack,
we find Germany surprised:
\begin{quote}
\small
Germany: Not really sure what to say, except that I regret you did what you did.
\end{quote}

Such scenarios suggest an important research challenge: is the forthcoming betrayal signaled by linguistic cues appearing in the (ostensibly friendly) conversation between the betrayer and the eventual victim?
A positive answer would suggest not only that the betrayer
unknowingly reveals their future treachery, but also that the eventual victim fails to notice these signals.
Capturing these signals computationally would therefore mean outperforming the human players.

In this work, we provide a framework for analyzing a dyad's evolving
communication patterns and provide evidence
of
 subtle but consistent
conversational patterns that foretell the unilateral dissolution of a friendship.
In particular, imminent betrayal is signaled by sudden changes in the balance of
conversational attributes such as positive sentiment, politeness, and
structured discourse.
Furthermore, we show that by exploiting these cues in a prediction setting we
can anticipate
imminent betrayal better than the human players.

After briefly describing the game (Section~\ref{sec:data}), we focus on how
the structure of the game provides convenient, reliable indicators of whether
pairs of participants are friends or foes (Section~\ref{sec:dyads}).  Given
these labels,
we explore linguistic features that are predictive of whether friendships will end in betrayal (Section~\ref{sec:longterm}) and, if so, when the
betrayal will happen (Section~\ref{sec:imminent}).

While our focus is on a single popular game, we choose methods that
generalize to other domains,
revealing dynamics present in other social
interactions (Section~\ref{sec:related}).  We discuss how
automatically predicting stable relationships and betrayal can more
broadly help advance the study of trust and relationships using
computational linguistics.


\section{Communication and Conflict in Diplomacy}\label{sec:data}

\begin{figure}[t]
  \begin{center}
  \includegraphics[width=.9\linewidth]{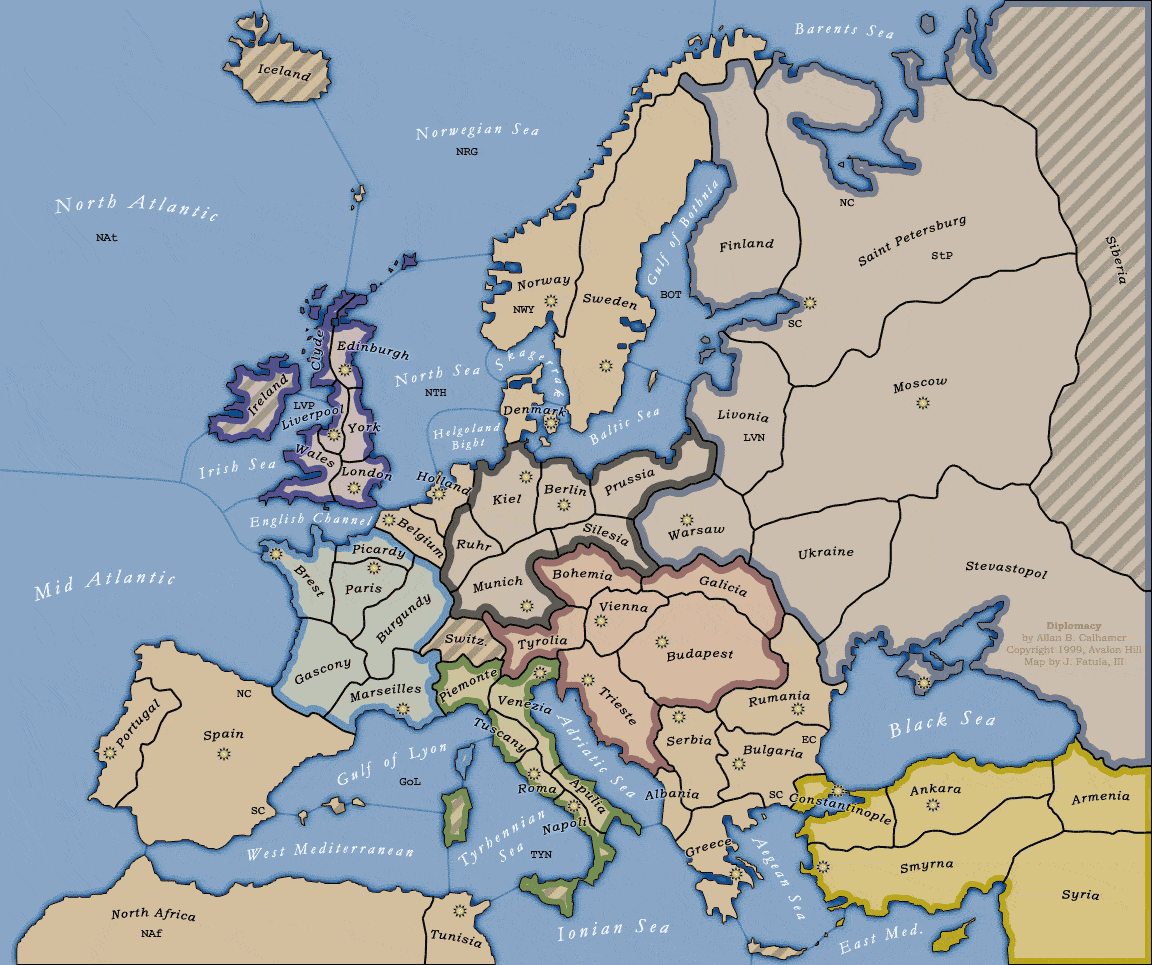}
  \end{center}
  \caption{The full Diplomacy board representing Europe circa 1914.
    The seven nations struggle to control the map.
  }
  \label{fig:board}
\end{figure}

A game of Diplomacy begins in
1901
with players casting themselves as the
European powers at the eve of the first world war: England, Germany, France,
Russia, Austria, Italy, and the Ottoman Empire.  The goal of the game (like
other war games such as Risk or Axis \& Allies) is to capture all of the
territories on the game board (Figure~\ref{fig:board}).  The games are divided
into years starting from
1901
and each year is divided into two seasons---Spring
and Fall.  Each season
consists of two
alternating phases:
\emph{diplomacy}---the players communicate
to form strategies---and
\emph{orders}---the players submit their moves for the season.
Seasons are therefore the main
unit
of
game
time.

\subsection{Movement, Orders, and Battles}
\label{sec:mechanics}

On the board, each player can operate a unit for each city they
control.  During each turn, these pieces have the option of moving to
an adjacent territory.  What makes Diplomacy unique is that all
players submit their written (or electronic) orders; these orders
are executed simultaneously; and
{\em there is no randomness}
(e.g., dice).  Thus, the
outcome of the game depends only on the communication, cooperation, and
movements of players.

When two units end their turn in the same territory, it implies a
battle.  Who wins the battle is
decided purely based
on numerical superiority
(ties go to defenders).  Instead of moving, a unit can support
another unit;
large armies can be created through intricate networks of support.
The side with the largest army wins the battle.

The process of \textit{supporting}
a unit is thus critical for
both
a successful offensive move and a successful defense.
  Often, a lone player lacks
the units to provide enough support to
his attacks
and thus needs
the help of
others.\footnote{
  While support can come from a player's own units, allies often combine resources.
  For example, if an English army in Belgium is attacking a
  Germany Army in Ruhr, a French army in Burgundy could
  strengthen that attack.
  This is accomplished by the French
  player
  submitting
  a move explicitly stating ``I support England's attack from
  Belgium to Ruhr''.}
 Because these orders (both movement and
support) are machine readable, we have a clear indication of when
players are working together (supporting each other) or working
against each other (attacking each other); we will use this to define
relationships between players (Section~\ref{sec:dyads}).  However,
coordinating these actions between players requires cooperation and
\emph{diplomacy}.

\subsection{Communication}

In the \emph{diplomacy} phase of the game, players talk to each
other.  These conversations are either
public
or---more
typically---one-on-one.  Conversations include greetings, extra-game
discussions (e.g., ``did you see Game of Thrones?''), low-level
tactics (``if you attack Armenia, I'll support you''), and high-level
strategy (``we need to control
Central
Europe'').
The content of these messages forms the object of our study.

Because of the centrality of language to Diplomacy, we can learn the rhetorical
and social devices players use to build and break
trust.
Because this language
is embedded in
every
game, it has convenient properties: similar situations are
repeated, the goals are clear, and machine-readable orders
confirm
which players are enemies and which are friends.
In the next section, we
explore the Diplomacy data.

\subsection{Preprocessing}

We use games from two popular online platforms for playing
Diplomacy.\footnote{Anonymized transcripts and more information available at
\url{http://vene.ro/betrayal/}
}
The average season of an online Diplomacy game lasts nine days.
We remove non-standard games
caused by differences between the two platforms, as well as
games that are still in progress.
Moreover, in each game, we filter out setup messages, regulatory messages to and from the
administrator of the game and messages declaring the state of the game,
keeping
only messages between the players.
This leaves
249
games with
145.000
total messages.

The dataset confirms that communication is an essential part of Diplomacy:
half of the games have over 515 messages exchanged between the players,
while the top quartile has over 750 messages per game.  Also, non-trivial
messages (with at least one sentence) tend to be complex: over half of them
have at least
five
sentences, and the top quartile consists of messages with
eight
or more sentences.


\section{Relationships and Their Stability}\label{sec:dyads}

\begin{figure}[t]
\scriptsize
  \centering
\rowcolors{2}{gray!25}{white}
\begin{tabular}{ccp{.65\linewidth}}
{\small Event} & {\small Time} & {\small What happened} \\
\hline
$F_1$ & 4 & \B supports \V's army in Vienna \\
$F_2$ & 3 & \V supports \B's attack from Warsaw to Silesia \\
$F_3$ & 3 & \B again supports \V in Vienna \\
$F_4$ & 1 & \V supports \B's move from Venice to Tyrolia \\
$H_5$ & 0 & \B attacks \V in Vienna \\
$H_6$ & -1 & \V retaliates, attacking \B in Warsaw \\
\hline
\end{tabular}
\includegraphics[width=0.35\textwidth]{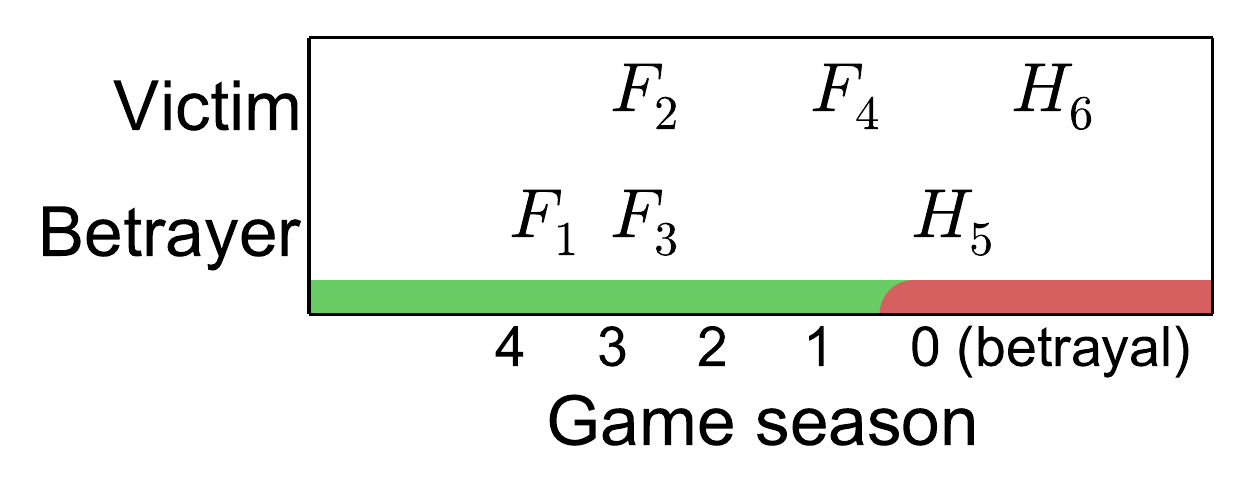}
\caption{
A friendship between Player~\B (eventual \betrayer) and Player~\V (eventual \victim) unravels.  For
the first four events, the players exchange \textbf{F}riendly acts (in green).  Eventually
\B's unilateral hostile act betrays \V's trust,
leading to hostility
 (in red).
 The dissolution takes place at the time of the first hostile act ($t=0$) and
 we index game seasons going back from the betrayal, such that lower
 indices mean betrayal is nearer.
}
\label{fig:sketch}
\end{figure}

In this section, we explore how interactions within the game of Diplomacy define
the relationships between players.
While such dyadic relationships can be undefined
(e.g., England and Turkey are in opposite corners of the map),
specific interactions between players
indicate
whether
they are friendly or
hostile to each other.

\xhdr{Friendships and hostilities} Alliances are a natural part of the game of
Diplomacy.  While the best outcome
for a player is a
{\em solo victory} against all other players, this
is rare and difficult to
achieve
 without any cooperation and assistance. Instead,
 the game's structure encourages players
 to form
 long-term alliances.
  Allies 
   often settle for (less prestigious)
  team victories, but these coalitions can also crumble as players seek a (more
  prestigious) solo victory for themselves.
  This game dynamic naturally leads to the formation of {\em friendly and
    hostile dyads}, which are relatively easy to identify
through
post-hoc analysis
  of the game, as explained next.

\xhdr{Acts of friendship}
Diplomacy provides a {\em support} option for players to help each other: this
game mechanism (discussed in Section~\ref{sec:data}) provides
unequivocal evidence of friendship.
When two players engage in a series of such friendly acts, we will say
that the two
are in a relation of \textit{friendship}.

\xhdr{Acts of hostility}
Unlike support, hostile actions are not explicitly marked in Diplomacy.
We consider two players to be hostile if they get involved in any unambiguous
belligerent action, such as invading one another's
territory, or
if one supports an enemy of the other.\footnote{
  In Diplomacy all game actions are simultaneous, and this can lead to ambiguous
  interpretation of the nature of a pair's
   interactions. Our definition
  of hostility intentionally discards such ambiguous evidence.  For instance, if
  two players attempt to move into the same unoccupied territory, this is not
  necessarily aggressive:
  allies sometimes use this tactic (``bouncing'') to ensure
  that a territory remains unoccupied.
}

\xhdr{Betrayal}
As in real life, friendships can be broken unilaterally: an individual can
\textit{betray}
his
 friend by engaging in a hostile act towards
her.
Figure~\ref{fig:sketch} shows two players who started out as friends (green) but
became hostile (red) after a betrayal. Importantly, until the last act of
friendship (game season $t=1$), 
the \textit{\victim} is unaware that she
will be betrayed
(otherwise she would not have engaged in an act of friendship)
and the
\textit{\betrayer} has no interest in signaling
his
planned duplicity
to his partner.

This setting poses the following research challenge: are there linguistic cues
that appear during the friendly conversations and portend
the
upcoming betrayal?  A positive answer would have two implications: the \betrayer
unknowingly hints at
his
future treachery, and the \victim
could have noticed it, but did not.
We will explore this question in the following sections.

\xhdr{Relationship stability}
Before venturing into the linguistic analysis of
betrayals, we briefly explore the dynamics underlying these state
transitions.
We find that, as in real life, friendships are much more likely to collapse
into hostilities than the reverse: in Diplomacy,
the probability of a friendship to
dissolve into enmity is about
five times greater than
that of hostile players
becoming friends.
The history of the relationship also matters.
A friendship built on the foundation of many cooperative acts is more likely to
endure than friendship with a short history, and
long-lasting conflict is less likely to become a friendship.
In numbers, the probability
that a
two season
long friendship
ends
is 35\%, while
for pairs who have helped each other for
ten
or more seasons, the probability
of betrayal is only 23\%.  Similarly, the probability
that a
two season
long
conflict
resolves
is 7\%, while players at war for over
ten
seasons have only
a 5\% chance to make up.
These numbers aren't particularly shocking---the idea that the passage
of time has an effect on the strength of a relationship is intuitive.
For the purposes of this study, we
control for
such
 effects
in order to capture purely linguistic hints of betrayal.

Starting from the relationship definitions discussed in this section, in what follows we show how
 subtle
linguistic patterns of in-game player conversations can reveal whether
or not
a
friendship
will turn hostile or not.




\section{Language Foretelling Betrayal}\label{sec:longterm}

In this section, we examine whether the conversations between two Diplomacy
allies contain linguistic cues foretelling
if
their friendship will last or
end in betrayal.  We
expect these cues to be subtle, since we only consider
messages exchanged when the two individuals are being ostensibly
friendly; when at least one of them---the eventual victim---is
unaware of the relationship's fate.

\subsection{What Constitutes a Betrayal}

To find betrayals, we must first find friendships.  Building on the
discussion from Section \ref{sec:dyads}, we consider a friendship to
be
\textit{stable} if it is
 ongoing, established, and reciprocal.  Thus, we focus on
relationships that contain at least two
consecutive and reciprocated
acts of friendships
that span at last at least three
seasons in game time.  We also
check that no more than five seasons pass between two acts of friendships, as
friendships can fade.

Betrayals are
established and reciprocal
friendships that
end
with at least two hostile
acts.  The person initiating the first of these hostile acts is the \textit{betrayer}, while the other person is the \textit{victim}.\footnote{
In rare cases, the betrayal can be mutual (i.e., both players start
attacking each other in the same season).
In such cases, we consider both betrayals.
}

For each
betrayal instance, we find the most similar stable friendship that was
\emph{never} dissolved by betrayal.  Using a greedy heuristic, we select friendships that match the betrayals on two
statistics: the length of the friendship and number of
seasons
since the start of
the game.
After this matching process, we find no significant difference in either of the two
variables (Mann-Whitney $p>0.3$).
Matching betrayals with \nobetrayals in this fashion removes historical and relationship-type effects such as those discussed in Section \ref{sec:dyads}, and focuses the comparison on the variable of interest:
whether a given stable friendship will end in a betrayal or not.

\subsection{Linguistic Harbingers of Betrayal}

\begin{figure*}[t]
\captionsetup[subfigure]{justification=centering}
\centering
\begin{subfigure}[t]{0.325\textwidth}
    \centering
    \includegraphics[width=\textwidth]{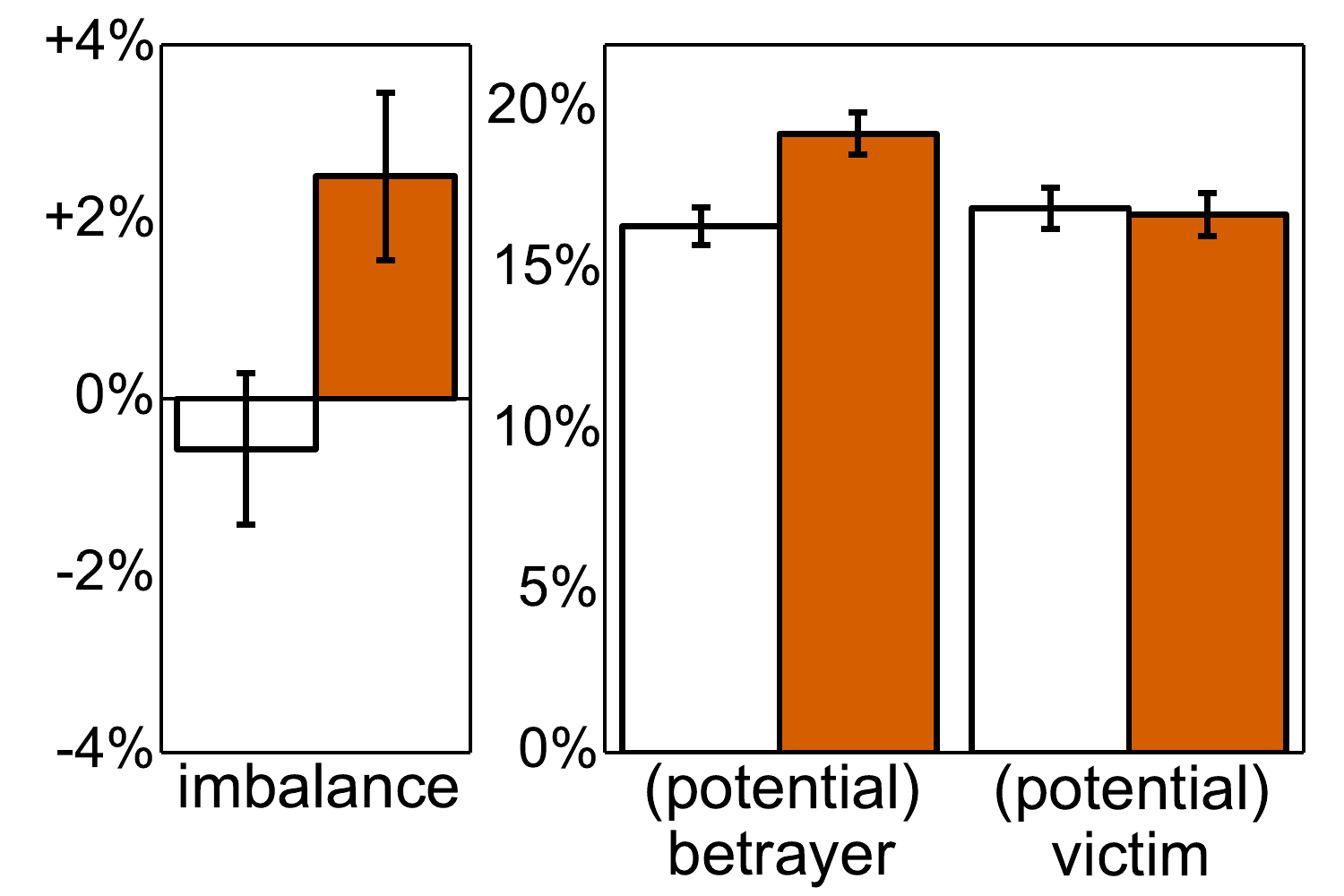}
    \caption{Positive sentiment
    \\(percentage of sentences)}
    \label{fig:sentiment}
\end{subfigure}
\begin{subfigure}[t]{0.325\textwidth}
    \centering
    \includegraphics[width=\textwidth]{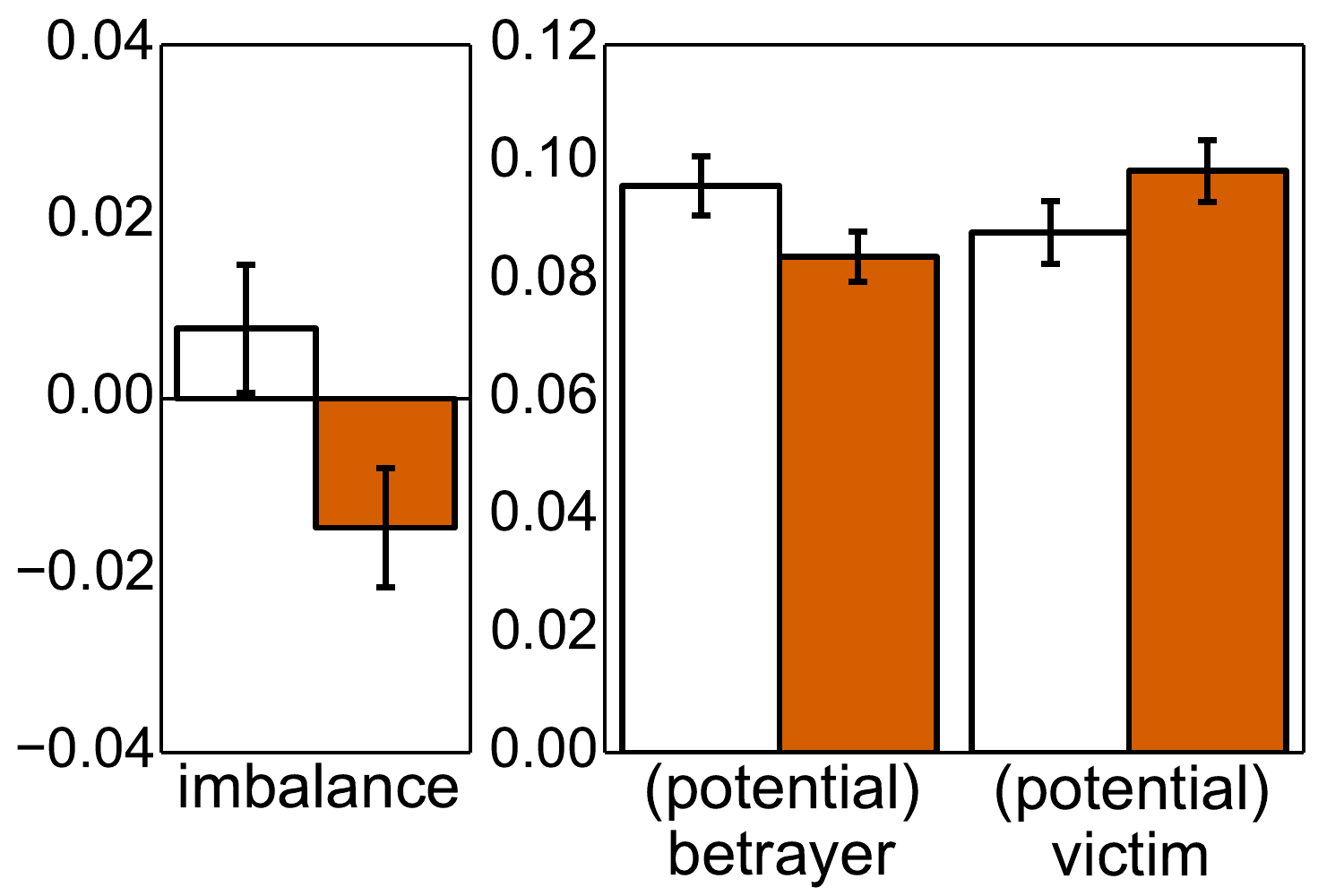}
    \caption{
    \Planning
    discourse markers
    \\(avg. number per sentence)}
    \label{fig:temporal}
\end{subfigure}
\begin{subfigure}[t]{0.325\textwidth}
    \centering
    \includegraphics[width=\textwidth]{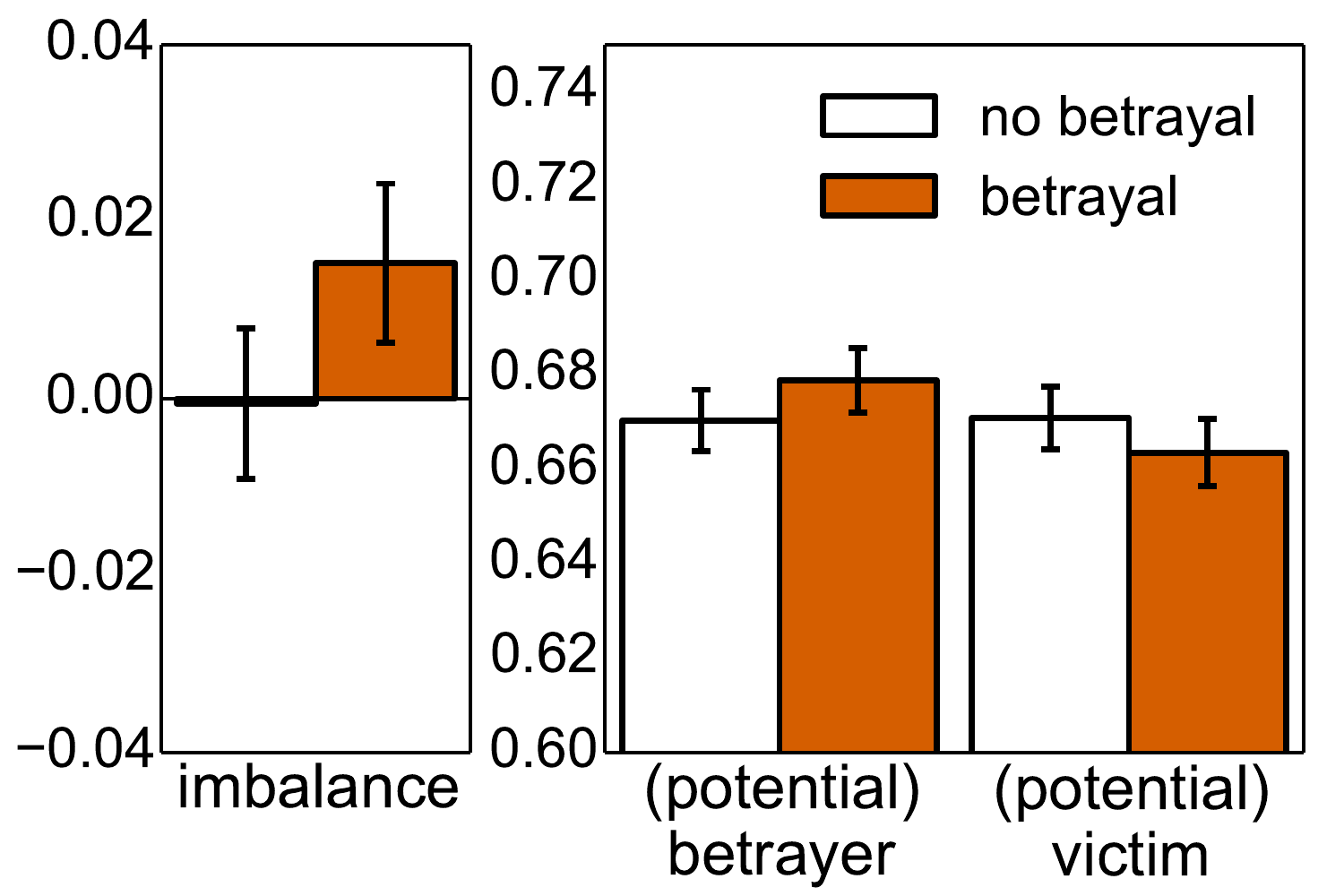}
    \caption{Politeness
    \\(avg. message score)}
    \label{fig:politeness}
\end{subfigure}
\caption{%
Friendships that will end in betrayal are imbalanced.
The
eventual betrayer is more positive, more polite, but plans less than
the victim.
The white bars correspond to
matched lasting friendships,
where the roles of
potential
betrayer and victim are arbitrarily assigned;
in these cases,
the imbalances disappear.
Error bars mark
bootstrapped standard errors
\protect\cite{efron1979bootstrap}.
}
\label{fig:features}
\end{figure*}

Now we switch to exploring linguistic features that correlate with
future betrayal in the controlled setting described above.  We start
from the intuition that a stable relationship should be balanced
\cite{Jung:2012:GHB:2207676.2208523}: friends will help each other
while enemies will fight each other.  A precarious friendship might
feel one-sided, while a conflict may turn to friendship through a
magnanimous olive branch.  Therefore, we focus our attention on
linguistic features that have the potential to signal an imbalance in
the communication patterns of the dyad.

To ensure that we are studying conversational patterns that occur \textit{only}
when the two individuals in the dyad are ostensibly being
friends, we only extract features from the messages exchanged before the last
act of friendship, that is, before
the season labeled 1
in Figure~\ref{fig:sketch}.  Considering the nature of this setting, we can only
hope for subtle linguistic cues: if there were salient linguistic
signals, then the victim would
notice and
preempt
the betrayal.  Instead, they are taken by surprise; the following
is a typical reaction of a player after having been betrayed
by a friend:

\begin{quote}
\small
Well that move was sour. I'm guessing France put you up to it, citing my
large growth. This was a pity, as I was willing to give you the lion's share of
centers in the west. [...] If you voiced your concerns I would have supported
you in most of the western centers. Unfortunately now you have jumped out of
the pan into the fire.
\end{quote}

\begin{table*}[t!]
\small
\centering
\begin{tabular}{p{3cm} p{10cm}}
\toprule
Feature & Example sentence from the data \\
\midrule
Positive sentiment & I will still be thrilled if it turns out you win this war.\\
Negative sentiment & It's not a great outcome, but still an OK one. \\
Neutral sentiment  & Do you concur with my assumption? \\
\sectionrule
Claim              & But {\bfseries I believe} that E/F have discarded him and so {\bfseries I think} he might bite. \\
Premise            & I put Italy out {\bfseries because} I wanted to work with you.\\
\sectionrule
Comparison         & We can trade centers {\bfseries as} {\bfseries much as} we like {\bfseries after} that. \\
Contingency        & He did not, {\bfseries thus} we are {\bfseries indeed} in fine shape to continue as planned. \\
Expansion          & Would you {\bfseries rather} see WAR-UKR, or GAL-UKR? \\
Temporal           & I think he can {\bfseries still} be effective
to help me take TUN {\bfseries while} you take ROM.\\
\Planning           & HOL should fall {\bfseries next} year, and {\bfseries then} MUN and KIE shortly {\bfseries thereafter}.\\
Number of requests & \\
\sectionrule
Politeness         & I wonder if you shouldn't try to support Italy into MAR ... What do you think?\\
\sectionrule
Subjectivity       & I'm {\bfseries just} {\bfseries curious} what you {\bfseries think}. \\
\sectionrule
Talkativeness      & \\
\bottomrule
\end{tabular}
\caption{Summary of the linguistic cues we consider.
}
\label{tab:features}
\end{table*}
\xhdr{Sentiment}
Changes in the sentiment
expressed in conversation can
reflect
emotional
responses, social affect, as well as
the status of the relationship as a whole
\cite{Gottman:JournalOfMarriageAndFamily:2000,Wang:ProceedingsOfTheAcl:}.
We
quantify the proportion of exchanged sentences that
transmit
positive, neutral
and negative sentiment using the Stanford Sentiment
Analyzer~\cite{socher2013recursive}.\footnote{We collapse the few examples classified as
{\em extreme positive} and {\em extreme negative} examples into {\em positive}
and {\em negative}, respectively.}  Example sentences with these features,
as well as all other features we consider, can be found in
Table~\ref{tab:features}.

We find that an imbalance in the amount of positive sentiment expressed by the
two individuals is a subtle sign that the relation will end in betrayal
(Figure~\ref{fig:sentiment}, left; one-sample t-test on the imbalance,
$p=0.008$).
When looking closer at who is the source of this imbalance
(Figure~\ref{fig:sentiment}, right), we find that that it is the eventual
betrayer that uses significantly \textit{more positive sentiment} than the
control counterpart in the matched friendship (two-sample t-test, $p=0.001$).
This is somewhat surprising, and we speculate that this is
the betrayer
overcompensating
for
his
forthcoming actions.

\xhdr{Argumentation and Discourse}
Structured discourse and well-made arguments are essential in
persuasion~\cite{cialdini-00,anand-11}.  To capture discourse complexity, we
measure
the average number of explicit discourse connectors per sentence
\cite{prasad2008penn}.\footnote{We remove the connectors that appear in
    over
    20\%
    of the messages ({\em and}, {\em for}, {\em but}, {\em if},
    {\em as}, {\em or}, and {\em so}).}
These markers belong to four coarse classes: {\em
comparison}, {\em contingency}, {\em expansive}, and {\em temporal}.  To capture
{\em \planning}, we group temporal markers that refer to the future
(e.g.,``next'', ``thereafter'') in a separate category.
To quantify the level of
argumentation,
we calculate
average number of claim and premise markers per sentence, as identified
by \newcite{stab2014identifying}.  We also measure the number of request
sentences in each message, as identified by
the heuristics in
the Stanford
Politeness classifier \cite{danescucomputational}.

The structure of the discourse offers clues to whether the friendship
will last.  For example, Figure~\ref{fig:temporal} shows that in
friendships
doomed to end in betrayal, the victim uses planning
discourse markers significantly more often than the betrayer (one-sample t-test
on the imbalance,
$p=0.03$),
who is likely to
be
aware that the cooperation has
no future.
(More argumentation and discourse
features will be discussed in the following sections.)

\xhdr{Politeness}
Pragmatic information can also be informative of the relation between two individuals; for example \newcite{danescucomputational}
 show that differences in
levels of politeness
 can echo differences in status
and power.  We measure the
politeness of each message using the Stanford Politeness
classifier
and find that friendships that end in betrayal show a slight imbalance between
the level of politeness used by the two individuals (one-sample t-test on the
imbalance, $p=0.09$)
and that in those cases the future victim is the one that is less polite.

\xhdr{Subjectivity}
We
explored
phrases expressing opinion, accusation, suspicion, and
speculation taken from an automatically collected lexicon
\cite{riloff2003learning}, but did not find significant differences between betrayals and control friendships.

\xhdr{Talkativeness}
Another conversational aspect is
the amount of communication flowing between the players, in each direction.
To quantify this, we simply use the number of messages sent,
the average number of sentences per message, and the average number of
words per sentence.  Abnormal communication patterns can indicate
a
relationship breakdown. For example, friendships that dissolve are
characterized by an imbalance in the number of messages exchanged between the
two players (one-sample t-test, $p<0.001$).

These results show that there are indeed subtle linguistic imbalance
signals that are indicative of an
forthcoming
betrayal, even in a setting in which the victim is not aware of the
impending
betrayal.

\subsection{Predictive Power}

To test whether
these linguistic cues
have any predictive power
and to explore how they interact, we turn to
a binary
classification setting in which we try to detect whether
a
player \V will be betrayed
by a player \B.
(We will call player \V the
potential victim and player \B the potential betrayer.)
Expert humans---the actual victims---performed poorly on this task
and were not able to tell that they will be betrayed:
by virtue of how the dataset is constructed, the performance of the human players is at chance level.

We use the same balanced dataset of matched betrayals and lasting friendships as before
and consider as classification instances all the seasons coming from
each of
the two classes (663 betrayal seasons and 712 from lasting friendships).
As features,  %
we use the cues described above and summarized in Table~\ref{tab:features},
differentiated by
source:
\V or \B.  We use
logistic regression after univariate feature selection.
The best setting for the model parameters\footnote{
We optimize the number of features selected, the
scoring function used
(ANOVA or $\chi^2$), whether to automatically reweigh the classes,
the
regularizer
($\ell_1$ or $\ell_2$), and the
value of the regularization parameter C between $10^{-12}$ and
$10^{12}$.}
is selected
via
5-fold
cross validation, ensuring that
instances from the same game are never found in both train and validation folds. The resulting model achieves a cross-validation accuracy of 57\%
and a Matthews correlation coefficient
of
0.14,
significantly above chance
(52\%
accuracy
and 0 Matthews correlation coefficient),
with 95\%
bootstrapped
confidence.
This indicates that,
unlike the actual players,
the classifier is able to exploit subtle linguistic signals that
surface in the conversation.\footnote{Since our focus is on
understanding linguistic aspects of betrayal, rather than on achieving
the best possible performance on this particular Diplomacy task, we do
not use game-specific information, such as the players' position on
the map, or
any
information not accessible to
both
players.
}

The selected features and their coefficients are
reported in Table~\ref{tab:chosen-t1}.
On top of the observations we previously made, the feature ranking reveals that
writing more sentences per message is more common when one will betray.
Discourse features also prove relevant:
more complex discourse
indicates a lower likelihood of
the player betraying
(e.g.,\ Figure~\ref{fig:temporal}).

Overall, the selected linguistic features capture a consistent signal that
characterizes people's language when they are
about to betray: they tend to plan less than their victims, use less
structure in their communication, and are
overly
positive.

\begin{table}[t]
\centering
\small
\begin{tabular}{l l l l}
\toprule
From & Positive feature & From & Negative feature \\
\midrule
B & Positive sentiment & B & Expansion \\
B & Sentences & B & Comparison \\
  &  & B & Contingency \\
  &  & B & No. Words \\
  &  & B & \Planning \\
  &  & B & Negative sentiment \\
\bottomrule
\end{tabular}

\caption{Selected
features for recognizing upcoming betrayal,
in decreasing order of the
absolute value of their coefficients. The {\em
From
} column
indicates whether
the message containing the feature was sent by
the potential {\bf B}etrayer or the potential {\bf V}ictim.
(In this case, only betrayer features were selected.)
Positive features indicate that a friendship is more likely to end in betrayal.
}
\label{tab:chosen-t1}
\end{table}


\section{Sudden yet Inevitable Betrayal}\label{sec:imminent}

\begin{figure*}[t]
\captionsetup[subfigure]{justification=centering}
\centering
\begin{subfigure}[t]{0.325\textwidth}
    \centering
    \includegraphics[width=\textwidth]{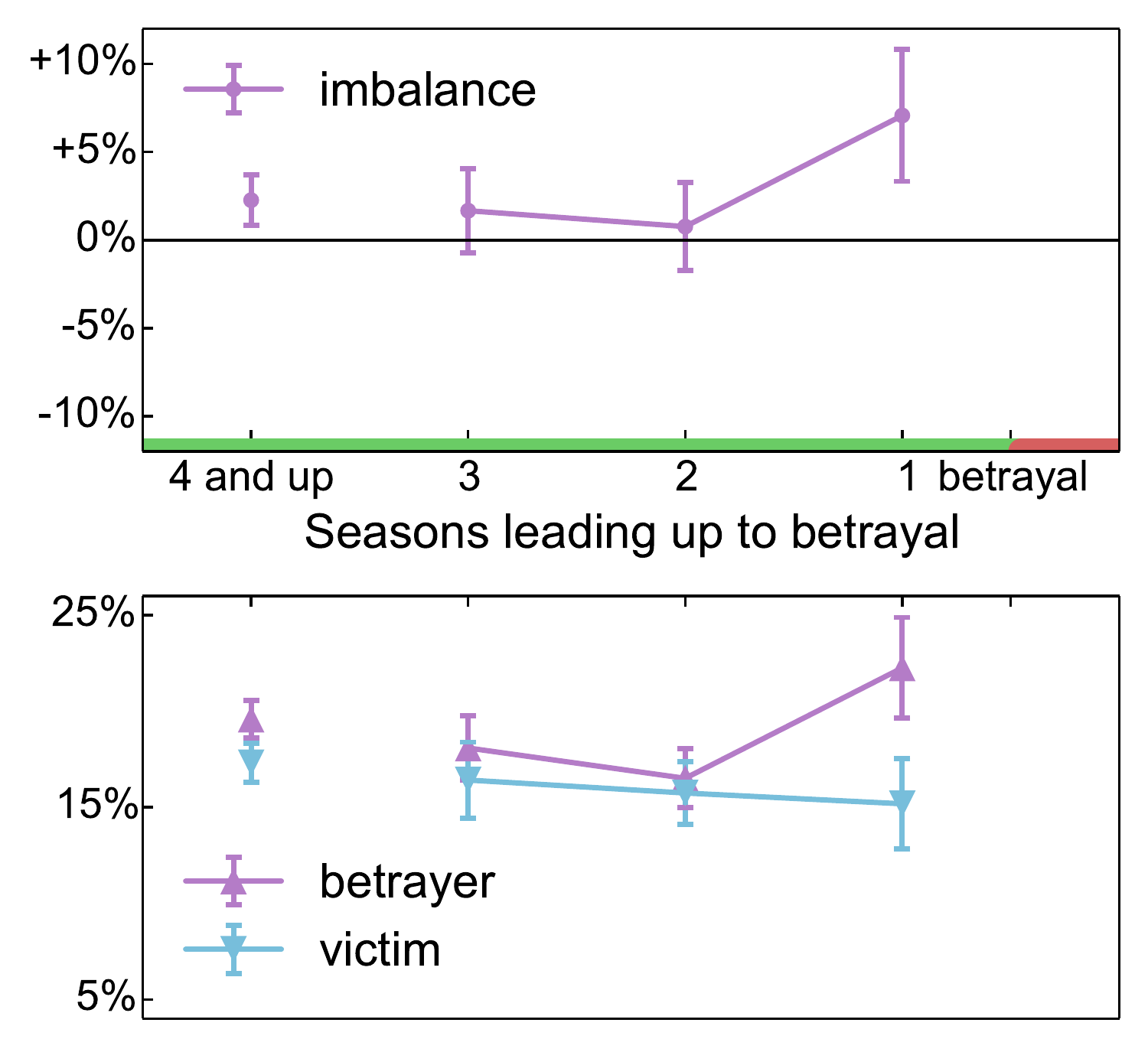}
    \caption{Positive sentiment
    \\(percentage of sentences)}
    \label{fig:sentiment_curve}
\end{subfigure}
\begin{subfigure}[t]{0.325\textwidth}
    \centering
    \includegraphics[width=\textwidth]{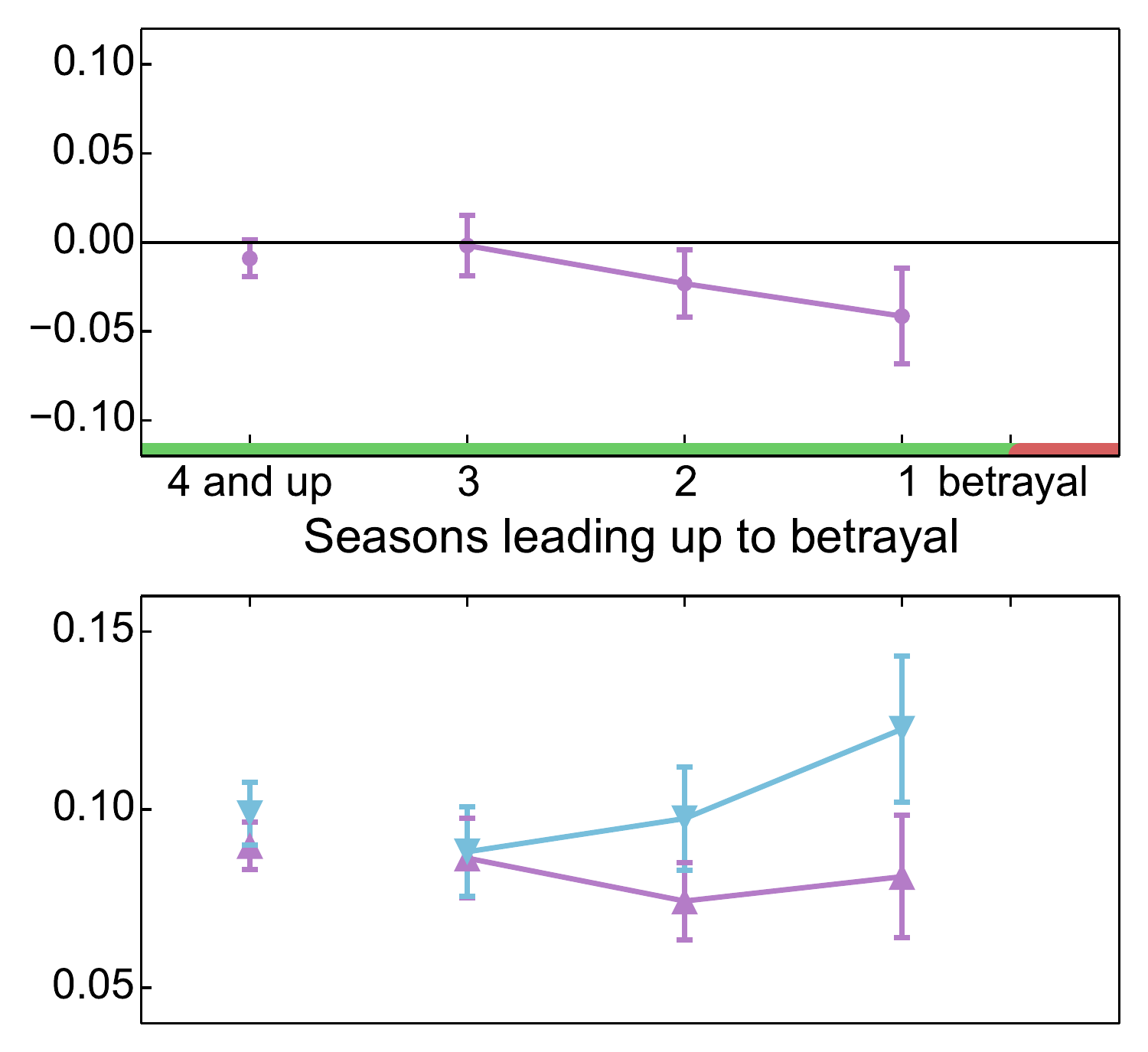}
    \caption{
    \Planning
    discourse markers
    \\(avg. number per sentence)}
    \label{fig:temporal_curve}
\end{subfigure}
\begin{subfigure}[t]{0.325\textwidth}
    \centering
    \includegraphics[width=\textwidth]{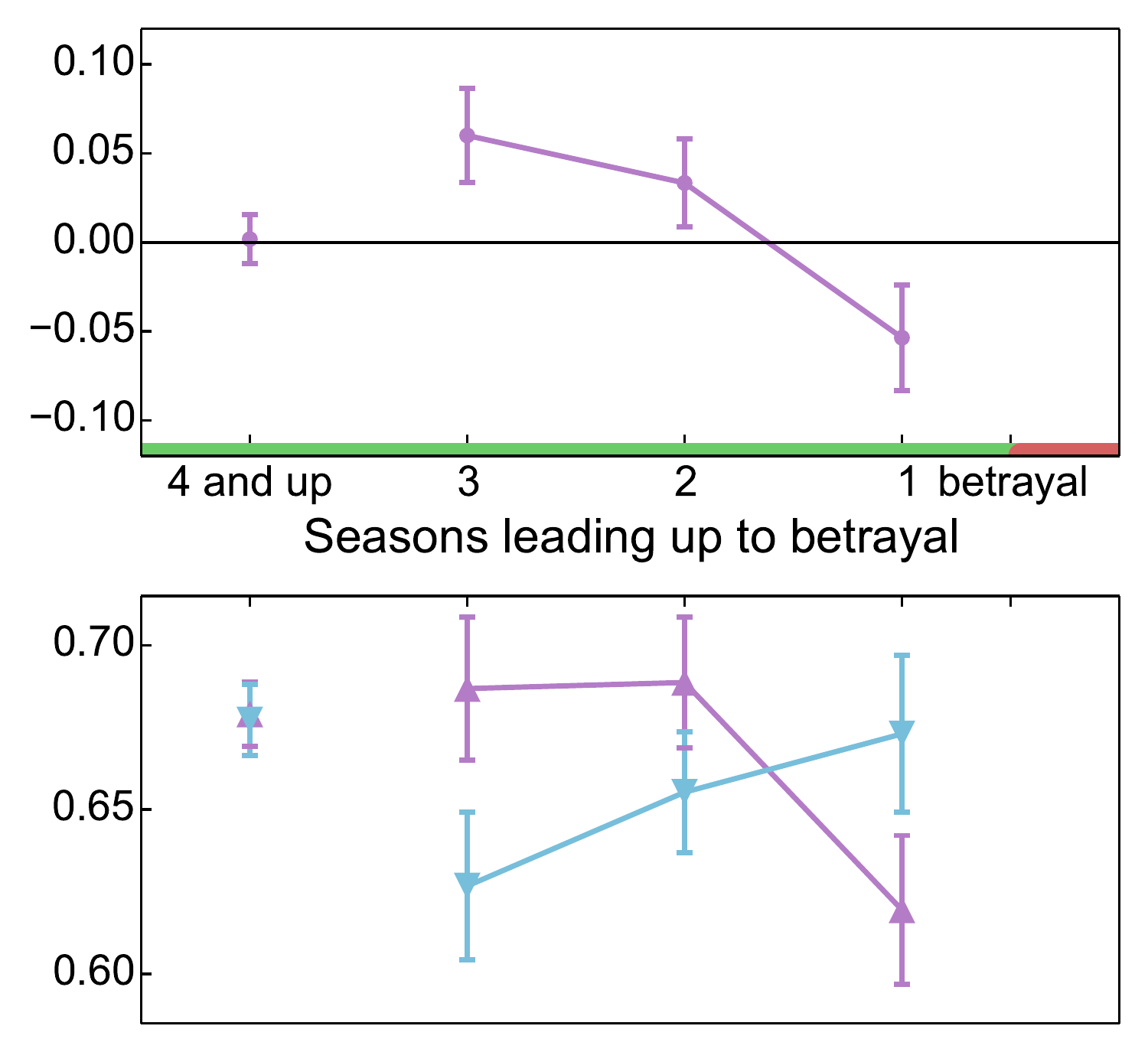}
    \caption{Politeness
    \\(avg. message score)}
    \label{fig:politeness_curve}
\end{subfigure}
\caption{
  Changes in balance can mark imminent betrayal.  As the breakdown approaches,
  the betrayer becomes more positive but less polite, and the victim tends to
  make more requests and become more polite.
  Error bars mark
  bootstrapped standard errors \protect\cite{efron1979bootstrap}.
}
\label{fig:curves}
\end{figure*}

The results from Section~\ref{sec:longterm} suggest that language cues can be
subtle signs of future relationship disruption.
Even though people are aware that most relationships eventually end, one would still prefer to reap their benefits as long as possible.
In Diplomacy,
despite
the common knowledge that
everyone prefers to win alone, players still take chances on long-lasting
alliances.
This leads to an alternate research question:
assuming that a relationship will be disrupted, how soon can one expect
to be betrayed?
This is still just as challenging for the expert human players,
as they were not able to anticipate and thereby avoid
betrayal.

 Next we
  investigate 
  if the variation of the linguistic cues over time
can predict
imminent change in the relationship.
We consider only the subset of
betrayals
 used in Section
\ref{sec:longterm}, and label each individual game season
with its distance from the end of the friendship (as in Figure \ref{fig:sketch}).
We prevent short alliances of circumstance from distorting the features close
to betrayal by keeping only
friendships lasting at least four seasons.

We consider the same
cues
 described in Table~\ref{tab:features},
and train a classifier to discriminate between the season preceding the
last friendly interaction
and
 all the
older seasons.
This learning task is imbalanced,
with only 14\% of the seasons being immediately before the betrayal.
Thus, we optimize
$F_1$ score and also measure the
Matthews correlation coefficient, which takes a value of 0 for uninformative
predictions (random or majority).
The best model achieves an $F_1$ score of 0.31 and a Matthews correlation
coefficient of 0.17,
significantly
better than chance
with 95\% bootstrapped confidence.
This shows that we can capture signs of imminent betrayal, something that
even the
skilled
human players have failed to do.
Furthermore, 39\% of the predicted false positives are within
two seasons of the last friendly act. This suggests that sometimes the
warning signs can appear slightly earlier.

The selected
features, displayed in
Table~\ref{tab:chosen-t2}, reflect some of the effects identified in
Section~\ref{sec:longterm}, such as the importance of positive sentiment
and
\planning discourse markers.
Betrayers have a tendency to use more positive sentiment
during the last moment of purported friendliness
(Figure~\ref{fig:sentiment_curve}).
Also, expressing more opinions through claims is a sign that one will not
betray right away.
Three of the discourse features (comparison, contingency and expansion)
are selected as {\em imbalance} features (they have near-opposite coefficients
for the betrayer and for the victim), indicating that
as
betrayal approaches,
victims are less eloquent than betrayers.
Interestingly,
some predictive signals come %
only
from the victim:
a partner using increasingly
more \planning words
is at higher risk of being betrayed
(Figure~\ref{fig:temporal_curve}). This
could
 be explained by the pressure that
making plans for
the future can put on a relationship. A similar reasoning applies
for making
many
requests.

We also find that a decrease in a partner's
politeness
presages
their imminent betrayal.
The change
in politeness over time (Figure~\ref{fig:politeness_curve})
reveals
a
reversal in the politeness imbalance of the pair. This explains why politeness
is not a good enough feature in detecting long-term betrayal. The behavior
could have
two intuitive explanations. On one hand, if the betrayer has planned the act in
advance, politeness can be a strategy for deception.  On the other hand, if the
betrayer receives impolite requests, the value of the relationship can decrease,
hastening
a betrayal.  We observe a similar dynamic for the average number of sentences
per message sent by the betrayer; the feature is selected in both
prediction tasks,
but
with opposite signs: more complex messages suggest that betrayal {\em will}
happen, but {\em not right away}.

Studying
language change
as betrayal
draws nearer
uncovers effects that cannot be seen when looking at an entire friendship on
average. For example, while excessively positive and polite partners are potential betrayers, people who have themselves suddenly become more polite are likely to become victims soon.

\begin{table}[t]
\centering
\small
\begin{tabular}{l l l l}
\toprule
From & Positive feature & From & Negative feature \\
\midrule
V & Comparison          & B & Claims\\
V & Positive sentiment  & B & Politeness \\
V & Contingency         & B & Contingency \\
V & \Planning           & B & Subjectivity \\
V & Requests            & B & Expansion \\
V & Expansion           & B & No. Sentences \\
  &                     & B & Comparison \\
\bottomrule
\end{tabular}
\caption{%
Selected features for recognizing imminent betrayal,
in decreasing order of the
absolute value of their coefficients.
The {\em
From
} column
indicates whether
the message containing the feature
comes from the
potential
{\bf B}etrayer or the
potential
{\bf V}ictim.
Positive features indicate that an exchange is more likely
to be followed by immediate betrayal.
}
\label{tab:chosen-t2}
\end{table}


\section{Relevance Beyond the Game}\label{sec:related}

While discovering betrayal in one online game is a fun and novel task,
our work connects with broader research in computational social
science.  In this section we describe how our work tackles issues that
previous research on alliances, negotiation, and relationships have
faced.

Cooperation and relationship building are an essential part of many
activities: completing a group project, opening a business, or forging
a new relationship.  Each of these has been the subject of extensive
research to understand what makes for effective relationships.
\newcite{Jung:2012:GHB:2207676.2208523} show that a balanced working
relationship is more likely to lead to better performance on tasks
like pair programming. \newcite{imai-10} show that understanding
cultural norms improves negotiations.  While these data are elicited
in the lab, our ``found'' data are inexpensive because Diplomacy games
are fun and inherently anonymized.

Romance is a popular and more real-world phenomenon that helps us understand how
relationships form and dissolve.
The research that tells us how language shapes early
dating~\cite{Ranganath:2009:YMD:1699510.1699554} and whether an
existing relationship will
continue~\cite{Slatcher01082006,Gottman:JournalOfMarriageAndFamily:2000,Ireland+al:2010a}
is formed from an incomplete sample of a course of a relationship.  In
contrast, a game of Diplomacy is shorter than almost any marriage and
we have a complete account of all interactions
throughout
the entire
relationship.
Furthermore, this work focuses on the unilateral and asymmetric act of betrayal, rather than on the question of whether a relation will last.


Playing Diplomacy online is less tangible than a romantic
relationship, but understanding trust and deception in online interactions \cite{riegelsberger-03,Newman:2003,hancock2007lying,Ott:2011,Feng:2012} is particularly important because the Internet marketplace is a
growing driver of economic growth~\cite{boyd-03}.  Diplomacy offers a setting in which deception occurs  spontaneously in the context of complex relationships.


\section{Conclusions}\label{sec:conclusion}

Despite people's best effort to hide it, the intention to betray can leak
through the language one uses.
Detecting it is
not a task that we expect to be solvable with high accuracy, as that
would entail a reliable ``recipe'' for avoiding betrayal in relationships; in this unrealistic scenario, betrayals would be unlikely to exist.
While the effects we find are subtle,
they bring new insights into the relation between linguistic balance and stability in relationships.

Although
we use one game to develop our methodology, the framework developed here can be extended to be applied to a wide range of social interaction.
Social
dynamics in collaborative settings can bear striking
similarities to those present in
war games.
For example, in Wikipedia
``edit wars''---where
attacks correspond to edit reverts---are common on issues relating
to politics, religion, history and nationality, among others
\cite{Kittur:ProceedingsOfChi:2007}.
 As in Diplomacy, Wikipedia editors form
alliances, argue and negotiate about possible compromises.  A challenge for future work is to find reliable linguistic cues that generalize well between such settings.


\section*{Acknowledgements}

This work is dedicated to all those who betrayed us.
We thank Mario Huys, Chris Babcock, and Christopher Martin for
providing the Diplomacy dataset.  We are grateful to Flavio Chierichetti, Malte Jung, Sendhil Mullainathan and the annonymous reviewers for their helpful comments. This work was conducted in part while Cristian Danescu-Niculescu-Mizil and Vlad Niculae were at the Max Planck Institute for Software Systems.
\mbox{Jordan Boyd-Graber} is supported by
\textsc{nsf} Grants \textsc{ccf}-1409287, \textsc{iis}-1320538, and
\textsc{ncse}-1422492.  Cristian Danescu-Niculescu-Mizil is supported by a Google Faculty Research Award.  Any opinions, findings, conclusions, or
recommendations expressed here are those of the authors and do not
necessarily reflect the view of the sponsor.

\bibliographystyle{style/acl}

\end{document}